# Nymble:
# a High-Performance Learning Name-finder

**Daniel M. Bikel**
BBN Corporation
70 Fawcett Street
Cambridge, MA 02138
dbikel@bbn.com

**Scott Miller**
BBN Corporation
70 Fawcett Street
Cambridge, MA 02138
szmiller@bbn.com

**Richard Schwartz**
BBN Corporation
70 Fawcett Street
Cambridge, MA 02138
schwartz@bbn.com

**Ralph Weischedel**
BBN Corporation
70 Fawcett Street
Cambridge, MA 02138
weisched@bbn.com

## Abstract
This paper presents a statistical, learned approach to finding names and other non-recursive entities in text (as per the MUC-6 definition of the NE task), using a variant of the standard hidden Markov model. We present our justification for the problem and our approach, a detailed discussion of the model itself and finally the successful results of this new approach.

## 1. Introduction

In the past decade, the speech recognition community has had huge successes in applying hidden Markov models, or HMM's to their problems. More recently, the natural language processing community has effectively employed these models for part-of-speech tagging, as in the seminal (Church, 1988) and other, more recent efforts (Weischedel et al., 1993). We would now propose that HMM's have successfully been applied to the problem of name-finding.

We have built a named-entity (NE) recognition system using a slightly-modified version of an HMM; we call our system "Nymble". To our knowledge, Nymble out-performs the best published results of any other learning name-finder. Furthermore, it performs at or above the 90% accuracy level, often considered "near-human performance".

The system arose from the NE task as specified in the last Message Understanding Conference (MUC), where organization names, person names, location names, times, dates, percentages and money amounts were to be delimited in text using SGML-markup. We will describe the various models employed, the methods for training these models and the method for "decoding" on test data (the term "decoding" borrowed from the speech recognition community, since one goal of traversing an HMM is to recover the hidden state sequence). To date, we have successfully trained and used the model on both English and Spanish, the latter for MET, the multi-lingual entity task.

## 2. Background

### 2.1 Name-finding as an Information-theoretic Problem

The basic premise of the approach is to consider the raw text encountered when decoding as though it had passed through a noisy channel, where it had been originally marked with named entities.[1] The job of the generative model is to model the original process that generated the name-class–annotated words, before they went through the noisy channel.

More formally, we must find the most likely sequence of name-classes ($NC$) given a sequence of words ($W$):

$$\Pr(NC \mid W) \qquad (2.1)$$

In order to treat this as a generative model (where it generates the original, name-class–annotated words), we use Bayes' Rule:

$$\Pr(NC \mid W) = \frac{\Pr(W, NC)}{\Pr(W)} \qquad (2.2)$$

and since the *a priori* probability of the word sequence—the denominator—is constant for any given sentence, we can maximize Equation 2.2 by maximizing the numerator alone.

---

[1] See (Cover and Thomas, 1991), ch. 2, for an excellent overview of the principles of information theory.

## 2.2 Previous Approaches to Name-finding

Previous approaches have typically used manually constructed finite state patterns (Weischedel, 1995, Appelt et al., 1995). For every new language and every new class of new information to spot, one has to write a new set of rules to cover the new language and to cover the new class of information. A finite-state pattern rule attempts to match against a sequence of tokens (words), in much the same way as a general regular expression matcher.

In addition to these finite-state pattern approaches, a variant of Brill rules has been applied to the problem, as outlined in (Aberdeen et al., 1995).

## 2.3 Interest in Problem and Potential Applications

The atomic elements of information extraction—indeed, of language as a whole—could be considered the who, where, when and how much in a sentence. A name-finder performs what is known as surface- or lightweight-parsing, delimiting sequences of tokens that answer these important questions. It can be used as the first step in a chain of processors: a next level of processing could relate two or more named entities, or perhaps even give semantics to that relationship using a verb. In this way, further processing could discover the "what" and "how" of a sentence or body of text.

Furthermore, name-finding can be useful in its own right: an Internet query system might use name-finding to construct more appropriately-formed queries: "When was Bill Gates born?" could yield the query `"Bill Gates"+born`. Also, name-finding can be directly employed for link analysis and other information retrieval problems.

## 3. Model

We will present the model twice, first in a conceptual and informal overview, then in a more-detailed, formal description of it as a type of HMM. The model bears resemblance to Scott Miller's novel work in the Air Traffic Information System (ATIS) task, as documented in (Miller et al., 1994).

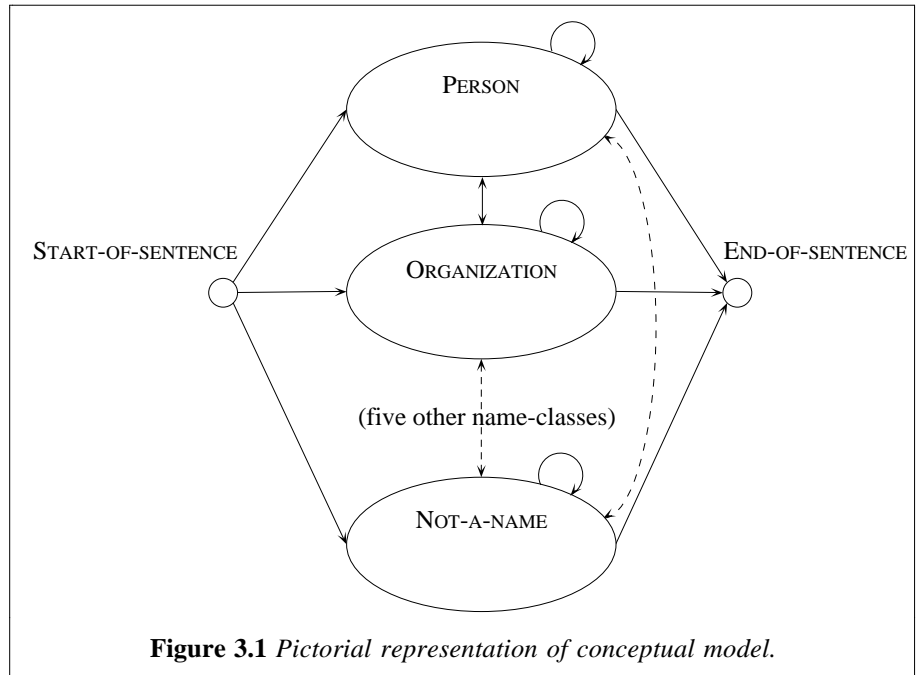

**Figure 3.1** *Pictorial representation of conceptual model.*

### 3.1 Conceptual Model

Figure 3.1 is a pictorial overview of our model.

Informally, we have an ergodic HMM with only eight internal states (the name classes, including the NOT-A-NAME class), with two special states, the START- and END-OF-SENTENCE states. Within each of the name-class states, we use a statistical bigram language model, with the usual one-word-per-state emission. This means that the number of states in each of the name-class states is equal to the vocabulary size, $|V|$.

The generation of words and name-classes proceeds in three steps:
1. Select a name-class *NC*, conditioning on the previous name-class and the previous word.
2. Generate the first word inside that name-class, conditioning on the current and previous name-classes.
3. Generate all subsequent words inside the current name-class, where each subsequent word is conditioned on its immediate predecessor.

These three steps are repeated until the entire observed word sequence is generated. Using the Viterbi algorithm, we efficiently search the entire space of all possible name-class assignments, maximizing the numerator of Equation 2.2, Pr(*W*, *NC*).

Informally, the construction of the model in this manner indicates that we view each type of "name" to be its own language, with separate bigram probabilities for generating its words. While the number of word-states within each name-class is equal to $|V|$, this "interior" bigram language model is ergodic, *i.e.*, there is a probability associated with every one of the $|V|^2$ transitions. As a parameterized, trained model, if such a transition were never observed, the model "backs off" to a less-powerful model, as described below, in §3.3.3 on p. 4.

| Word Feature | Example Text | Intuition |
|---|---|---|
| `twoDigitNum` | 90 | Two-digit year |
| `fourDigitNum` | 1990 | Four digit year |
| `containsDigitAndAlpha` | A8956-67 | Product code |
| `containsDigitAndDash` | 09-96 | Date |
| `containsDigitAndSlash` | 11/9/89 | Date |
| `containsDigitAndComma` | 23,000.00 | Monetary amount |
| `containsDigitAndPeriod` | 1.00 | Monetary amount, percentage |
| `otherNum` | 456789 | Other number |
| `allCaps` | BBN | Organization |
| `capPeriod` | M. | Person name initial |
| `firstWord` | *first word of sentence* | No useful capitalization information |
| `initCap` | Sally | Capitalized word |
| `lowerCase` | can | Uncapitalized word |
| `other` | , | Punctuation marks, all other words |

**Table 3.1** *Word features, examples and intuition behind them*

### 3.2 Words and Word-Features

Throughout most of the model, we consider words to be ordered pairs (or two-element vectors), composed of word and word-feature, denoted $\langle w, f \rangle$. The word feature is a simple, deterministic computation performed on each word as it is added to or looked up in the vocabulary. It produces one of the fourteen values in Table 3.1.

These values are computed in the order listed, so that in the case of non-disjoint feature-classes, such as `containsDigitAndAlpha` and `containsDigitAndDash`, the former will take precedence. The first eight features arise from the need to distinguish and annotate monetary amounts, percentages, times and dates. The rest of the features distinguish types of capitalization and all other words (such as punctuation marks, which are separate tokens). In particular, the `firstWord` feature arises from the fact that if a word is capitalized and is the first word of the sentence, we have no good information as to why it is capitalized (but note that `allCaps` and `capPeriod` are computed before `firstWord`, and therefore take precedence).

The word feature is the one part of this model which is language-dependent. Fortunately, the word feature computation is an extremely small part of the implementation, at roughly ten lines of code. Also, most of the word features are used to distinguish types of numbers, which *are* language-independent.[2] The rationale for having such features is clear: in Roman languages, capitalization gives good evidence of names.[3]

### 3.3 Formal Model

This section describes the model formally, discussing the transition probabilities to the word-states, which "generate" the words of each name-class.

#### 3.3.1 Top Level Model

As with most trained, probabilistic models, we have a most accurate, most powerful model, which will "back off" to a less-powerful model when there is insufficient training, and ultimately back-off to unigram probabilities.

In order to generate the first word, we must make a transition from one name-class to another, as well as calculate the likelihood of that word. Our intuition was that a word preceding the start of a name-class (such as "Mr.", "President" or other titles preceding the PERSON name-class) and the word following a name-class would be strong indicators of the subsequent and preceding name-classes, respectively. Accordingly, the probability for generating the first word of a name-class is factored into two parts:

$$\Pr(NC \mid NC_{-1}, w_{-1}) \cdot \Pr(\langle w, f \rangle_{first} \mid NC, NC_{-1}).$$

(3.1)

---

[2] Non-English languages tend to use the comma and period in the reverse way in which English does, *i.e.*, the comma is a decimal point and the period separates groups of three digits in large numbers. However, the re-ordering of the precedence of the two relevant word-features had little effect when decoding Spanish, so they were left as is.

[3] Although Spanish has many lower-case words in organization names. See §4.1 on p. 6 for more details.

The top level model for generating all but the first word in a name-class is

$$\Pr(\langle w,f\rangle \mid \langle w,f\rangle_{-1}, NC). \quad (3.2)$$

There is also a magical "+end+" word, so that the probability may be computed for any current word to be the final word of its name-class, i.e.,

$$\Pr(\langle +end+, \text{other}\rangle \mid \langle w,f\rangle_{final}, NC). \quad (3.3)$$

As one might imagine, it would be useless to have the first factor in Equation 3.1 be conditioned off of the +end+ word, so the probability is conditioned on the previous *real* word of the previous name-class, i.e., we compute

$$\Pr(NC \mid NC_{-1}, w_{-1}) \begin{cases} w_{-1} = +end+ \text{ if} \\ \quad NC_{-1} = \text{START-OF-SENTENCE} \\ w_{-1} = \text{last observed word otherwise} \end{cases} \quad (3.4)$$

Note that the above probability is not conditioned on the word-feature of $w_{-1}$, the intuition of which is that in the cases where the previous word would help the model predict the next name-class, the word feature—capitalization in particular—is not important: "Mr." is a good indicator of the next word beginning the PERSON name-class, regardless of capitalization, especially since it is almost never seen as "mr.".

### 3.3.2 Calculation of Probabilities

The calculation of the above probabilities is straightforward, using events/sample-size:

$$\Pr(NC \mid NC_{-1}, w_{-1}) = \frac{c(NC, NC_{-1}, w_{-1})}{c(NC_{-1}, w_{-1})} \quad (3.5)$$

$$\Pr(\langle w,f\rangle_{first} \mid NC, NC_{-1}) = \frac{c(\langle w,f\rangle_{first}, NC, NC_{-1})}{c(NC, NC_{-1})} \quad (3.6)$$

$$\Pr(\langle w,f\rangle \mid \langle w,f\rangle_{-1}, NC) = \frac{c(\langle w,f\rangle, \langle w,f\rangle_{-1}, NC)}{c(\langle w,f\rangle_{-1}, NC)} \quad (3.7)$$

where $c()$ represents the number of times the events occurred in the training data (the *count*).

### 3.3.3 Back-off Models and Smoothing

Ideally, we would have sufficient training (or at least one observation of!) every event whose conditional probability we wish to calculate. Also, ideally, we would have sufficient samples of that upon which each conditional probability is conditioned, e.g., for $\Pr(NC \mid NC_{-1}, w_{-1})$, we would like to have seen sufficient numbers of $NC_{-1}$, $w_{-1}$. Unfortunately, there is rarely enough training data to compute accurate probabilities when "decoding" on new data.

#### 3.3.3.1 Unknown Words

The vocabulary of the system is built as it trains. Necessarily, then, the system knows about all words for which it stores bigram counts in order to compute the probabilities in Equations 3.1 – 3.3. The question arises how the system should deal with unknown words, since there are three ways in which they can appear in a bigram: as the current word, as the previous word or as both. A good answer is to train a separate, unknown word–model off of held-out data, to gather statistics of unknown words occurring in the midst of known words.

Typically, one holds out 10–20% of one's training for smoothing or unknown word–training. In order to overcome the limitations of a small amount of training data—particularly in Spanish—we hold out 50% of our data to train the unknown word–model (the vocabulary is built up on the first 50%), save these counts in training data file, then hold out the other 50% and concatenate these bigram counts with the first unknown word–training file. This way, we can gather likelihoods of an unknown word appearing in the bigram using all available training data. This approach is perfectly valid, as we are trying to estimate that which we have not legitimately seen in training. When decoding, if either word of the bigram is unknown, the model used to estimate the probabilities of Equations 3.1–3 is the unknown word model, otherwise it is the model from the normal training. The unknown word–model can be viewed as a first level of back-off, therefore, since it is used as a backup model when an unknown word is encountered, and is necessarily not as accurate as the bigram model formed from the actual training.

#### 3.3.3.2 Further Back-off Models and Smoothing

Whether a bigram contains an unknown word or not, it is possible that either model may not have seen this bigram, in which case the model backs off to a less-powerful, less-descriptive model. Table 3.2 shows a graphic illustration of the back-off scheme:

| Name-class Bigrams | First-word Bigrams | Non–first-word Bigrams |
|---|---|---|
| $\Pr(NC \mid NC_{-1}, w_{-1})$ | $\Pr(\langle w, f \rangle_{first} \mid NC, NC_{-1})$ | $\Pr(\langle w, f \rangle \mid \langle w, f \rangle_{-1}, NC)$ |
| $\vdots$ | $\vdots$ | $\vdots$ |
| $\Pr(NC \mid NC_{-1})$ | $\Pr(\langle w, f \rangle \mid \langle +begin+, \text{other} \rangle, NC)$ | $\Pr(\langle w, f \rangle \mid NC)$ |
| $\vdots$ | $\vdots$ | $\vdots$ |
| $\Pr(NC)$ | $\Pr(\langle w, f \rangle \mid NC)$ | $\Pr(w \mid NC) \cdot \Pr(f \mid NC)$ |
| $\vdots$ | $\vdots$ | $\vdots$ |
| $\dfrac{1}{\text{number of name - classes}}$ | $\Pr(w \mid NC) \cdot \Pr(f \mid NC)$ | $\dfrac{1}{\lvert V \rvert} \cdot \dfrac{1}{\text{number of word features}}$ |
| | $\vdots$ | |
| | $\dfrac{1}{\lvert V \rvert} \cdot \dfrac{1}{\text{number of word features}}$ | |

**Table 3.2** *Back-off strategy*

The weight for each back-off model is computed on-the-fly, using the following formula:

If computing Pr(X|Y), assign weight of $\lambda$ to the direct computation (using one of the formulae of §3.3.2) and a weight of $(1 - \lambda)$ to the back-off model, where

$$\lambda = \left(1 - \frac{\text{old } c(Y)}{c(Y)}\right) \frac{1}{1 + \dfrac{\text{unique outcomes of } Y}{c(Y)}}, \quad (3.8)$$

where "old $c(Y)$" is the sample size of the model from which we are backing off. This is a rather simple method of smoothing, which tends to work well when there are only three or four levels of back-off.[4] This method also overcomes the problem when a back-off model has roughly the same amount of training as the current model, via the first factor of Equation 3.8, which essentially ignores the back-off model and puts all the weight on the primary model, in such an equi-trained situation.

As an example—disregarding the first factor—if we saw the bigram "come hither" once in training and we saw "come here" three times, and nowhere else did we see the word "come" in the NOT-A-NAME class, when computing

Pr("hither" | "come", NOT-A-NAME),

we would back off to the unigram probability

Pr("hither" | NOT-A-NAME)

with a weight of $\frac{1}{3}$, since the number of unique outcomes for the word-state for "come" would be two, and the total number of times "come" had been the preceding word in a bigram would be four (a $1/(1+\frac{2}{4}) = \frac{2}{3}$ weight for the bigram probability, a $1 - \frac{2}{3} = \frac{1}{3}$ weight for the back-off model).

### 3.4 Comparison with a traditional HMM

Unlike a traditional HMM, the probability of generating a particular word is 1 for each word-state inside each of the name-class states. An alternative—and more traditional—model would have a small number of states within each name-class, each having, perhaps, some semantic significance, *e.g.*, three states in the PERSON name-class, representing a first, middle and last name, where each of these three states would have some probability associated with emitting any word from the vocabulary. We chose to use a bigram language model because, while less semantically appealing, such *n*-gram language models work remarkably well in practice. Also, as a first research attempt, an *n*-gram model captures the most general significance of the words in each name-class, without presupposing any specifics of the structure of names, á la the PERSON name-class example, above. More important, either approach is mathematically valid, as long as all transitions out of a given state sum to one.

### 3.5 Decoding

All of this modeling would be for naught were it not for the existence of an efficient algorithm for finding the optimal state sequence, thereby "decoding" the original sequence of name-classes. The number of possible state sequences for *N* states in an ergodic model for a sentence of *m* words is $N^m$, but, using dynamic programming and an appropriate merging of multiple theories when they converge on a particular state—the Viterbi decoding algorithm—a sentence can be "decoded" in time linear to the number of tokens in the sentence, O(*m*) (Viterbi, 1967). Since we are interested in recovering the name-class state sequence, we pursue eight theories at every given step of the algorithm.

---

[4] Any more levels of back-off might require a more sophisticated smoothing technique, such as deleted interpolation. No matter what smoothing technique is used, one must remember that smoothing is the art of estimating the probability of that which is unknown (*i.e.*, not seen in training).

## 4. Implementation and Results

### 4.1 Development History

Initially, the word-feature was not in the model; instead the system relied on a third-level back-off part-of-speech tag, which in turn was computed by our stochastic part-of-speech tagger. The tags were taken at face value: there were not *k*-best tags; the system treated the part-of-speech tagger as a "black box". Although the part-of-speech tagger used capitalization to help it determine proper-noun tags, this feature was only implicit in the model, and then only after two levels of back-off! Also, the capitalization of a word was submerged in the muddiness of part-of-speech tags, which can "smear" the capitalization probability mass over several tags. Because it seemed that capitalization would be a good name-predicting feature, and that it should appear earlier in the model, we eliminated the reliance on part-of-speech altogether, and opted for the more direct, word-feature model described above, in §3. Originally, we had a very small number of features, indicating whether the word was a number, the first word of a sentence, all uppercase, initial-capitalized or lower-case. We then expanded the feature set to its current state in order to capture more subtleties related mostly to numbers; due to increased performance (although not entirely dramatic) on every test, we kept the enlarged feature set.

Contrary to our expectations (which were based on our experience with English), Spanish contained many examples of lower-case words in organization and location names. For example, *departamento* ("Department") could often start an organization name, and adjectival place-names, such as *coreana* ("Korean") could appear in locations and by convention are not capitalized.

### 4.2 Current Implementation

The entire system is implemented in C++, atop a "home-brewed", general-purpose class library, providing a rapid code-compile-train-test cycle. In fact, many NLP systems suffer from a lack of software and computer-science engineering effort: run-time efficiency is key to performing numerous experiments, which, in turn, is key to improving performance. A system may have excellent performance on a given task, but if it takes long to compile and/or run on test data, the rate of improvement of that system will be miniscule compared to that which can run very efficiently. On a Sparc20 or SGI Indy with an appropriate amount of RAM, Nymble can compile in 10 minutes, train in 5 minutes and run at 6MB/hr. There were days in which we had as much as a 15% reduction in error rate, to borrow the performance measure used by the speech community, where error rate = 100% – F-measure. (See §4.3 for the definition of F-measure.)

### 4.3 Results of evaluation

In this section we report the results of evaluating the final version of the learning software. We report the results for English and for Spanish and then the results of a set of experiments to determine the impact of the training set size on the algorithm's performance in both English and Spanish.

For each language, we have a held-out development test set and a held-out, blind test set. We only report results on the blind test set for each respective language.

#### 4.3.1 F-measure

The scoring program measures both precision and recall, terms borrowed from the information-retrieval community, where

$$P = \frac{\text{number of correct responses}}{\text{number responses}} \text{ and}$$
$$R = \frac{\text{number of correct responses}}{\text{number correct in key}}. \quad (4.1)$$

Put informally, recall measures the number of "hits" vs. the number of possible correct answers as specified in the key file, whereas precision measures how many answers were correct ones compared to the number of answers delivered. These two measures of performance combine to form one measure of performance, the F-measure, which is computed by the weighted harmonic mean of precision and recall:

$$F = \frac{(\beta^2 + 1)RP}{(\beta^2 R) + P} \quad (4.2)$$

where $\beta^2$ represents the relative weight of recall to precision (and typically has the value 1). To our knowledge, our learned name-finding system has achieved a higher F-measure than any other learned system when compared to state-of-the-art manual (rule-based) systems on similar data.

#### 4.3.2 English and Spanish Results

Our test set of English data for reporting results is that of the MUC-6 test set, a collection of 30 WSJ documents (we used a different test set during development). Our Spanish test set is that used for MET, comprised of articles from the news agency AFP. Table 4.1 illustrates Nymble's performance as compared to the best reported scores for each category.

| Case | Language | Best Reported Score | Nymble |
|------|----------|---------------------|--------|
| Mixed | English | 96 | **93** |
| Upper | English | 89 | **91** |
| Mixed | Spanish | 93 | **90** |

**Table 4.1** *F-measure Scores*

### 4.3.3 The Amount of Training Data Required

With any learning technique, one of the important questions is how much training data is required to get acceptable performance. More generally how does performance vary as the training set size is increased or decreased? We ran a sequence of experiments in English and in Spanish to try to answer this question for the final model that was implemented.

For English, there were 450,000 words of training data. By that we mean that the text of the document itself (including headlines but not including SGML tags) was 450,000 words long. Given this maximum size of training available to us, we successfully divided the training material in half until we were using only one eighth of the original training set size or a training set of 50,000 words for the smallest experiment. To give a sense of the size of 450,000 words, that is roughly half the length of one edition of the Wall Street Journal.

The results are shown in a histogram in Figure 4.1 below. The positive outcome of the experiment is that half as much training data would have given almost equivalent performance. Had we used only one quarter of the data or approximately 100,000 words, performance would have degraded slightly, only about 1–2 percent. Reducing the training set size to 50,000 words would have had a more significant decrease in the performance of the system; however, the performance is still impressive even with such a small training set.

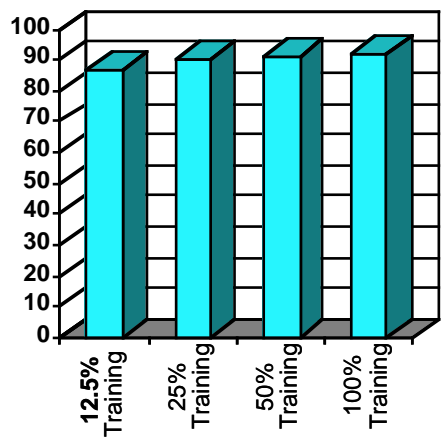

**Figure 4.1: Impact of Various Training Set Sizes on Performance in English.** *The learning algorithm performs remarkable well, nearly comparable to handcrafted systems with as little as 100,000 words of training data.*

On the other hand, the result also shows that merely annotating more data will not yield dramatic improvement in the performance. With increased training data it would be possible to use even more detailed models that require more data and could achieve significantly improved overall system performance with those more detailed models.

For Spanish we had only 223,000 words of training data. We also measured the performance of the system with half the training data or slightly more than 100,000 words of text. Figure 4.2 shows the results. There is almost no change in performance by using as little as 100,000 words of training data.

Therefore the results in both languages were comparable. As little as 100,000 words of training data produces performance nearly comparable to handcrafted systems.

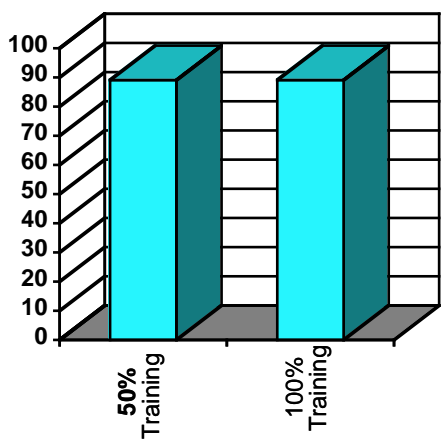

**Figure 4.2:** *Impact of Training Set Size on Performance in Spanish*

## 5. Further Work

While our initial results have been quite favorable, there is still much that can be done potentially to improve performance and completely close the gap between learned and rule-based name-finding systems. We would like to incorporate the following into the current model:

- lists of organizations, person names and locations
- an aliasing algorithm, which dynamically updates the model (where *e.g.* IBM is an alias of International Business Machines)
- longer-distance information, to find names not captured by our bigram model

## 6. Conclusions

We have shown that using a fairly simple probabilistic model, finding names and other numerical entities as specified by the MUC tasks can be performed with "near-human performance", often likened to an F of 90 or above. We have also shown that such a system can be trained efficiently and that, given appropriately and consistently marked answer keys, it can be trained on languages foreign to the trainer of the system; for example, we do not speak Spanish, but trained Nymble on answer keys marked by native speakers. None of the formalisms or techniques presented in this paper is new; rather, the approach to this task—*the model itself*—is wherein lies the novelty. Given the incredibly difficult nature of many NLP tasks, this example of a learned, stochastic approach to name-finding lends credence to the argument that the NLP community ought to push these approaches, to find the limit of phenomena that may be captured by probabilistic, finite-state methods.

## 8. Acknowledgements

The work reported here was supported in part by the Defense Advanced Research Projects Agency; a technical agent for part of the work was Fort Huachucha under contract number DABT63-94-C-0062. The views and conclusions contained in this document are those of the authors and should not be interpreted as necessarily representing the official policies, either expressed or implied, of the Defense Advanced Research Projects Agency or the United States Government.

We would also like to give special acknowledgement to Stuart Shieber, McKay Professor of Computer Science at Harvard University, who endorsed and helped foster the completion of this, the first phase of Nymble's development.